\documentclass[twoside,11pt]{article}

\usepackage{lipsum}
\usepackage{amsthm,amsmath,amsfonts,bm,dsfont}
\usepackage{amssymb}
\def\1{{\mathds{1}}}

\def\va{{\bm{a}}}

\def\vc{{\bm{c}}}

\def\vf{{\boldsymbol{f}}}
\def\vg{{\bm{g}}}

\def\vx{{\boldsymbol{x}}}
\def\vy{{\bm{y}}}

\def\mI{{\bm{I}}}

\def\mX{{\bm{X}}}

\DeclareMathAlphabet{\mathsfit}{\encodingdefault}{\sfdefault}{m}{sl}
\SetMathAlphabet{\mathsfit}{bold}{\encodingdefault}{\sfdefault}{bx}{n}

\newcommand{\E}{\mathbb{E}}

\newcommand{\R}{\mathbb{R}}

\newcommand{\Var}{\mathrm{Var}}

\newcommand{\Cov}{\mathrm{Cov}}

\DeclareMathOperator*{\argmax}{arg\,max}
\DeclareMathOperator*{\argmin}{arg\,min}

\usepackage{blindtext}
\usepackage[abbrvbib, preprint]{ArXiv}
\usepackage[svgnames]{xcolor}
\hypersetup{
hidelinks,
colorlinks=true,
linkcolor=Indigo,
urlcolor=DeepSkyBlue4,
citecolor=Indigo
}
\usepackage{todonotes}
\usepackage{amsmath} 
\usepackage{amssymb}
\usepackage{enumitem}
\usepackage{booktabs}
\usepackage{algorithm}
\usepackage{algpseudocode}

\usepackage{bm}
\usepackage{subfigure}
\usepackage{xcolor}
\usepackage{setspace}

\usepackage{wrapfig}
\usepackage{dsfont}
\def\1{{\mathds{1}}}
\DeclareFixedFont{\myfont}{OT1}{ptm}{m}{n}{8pt}
\DeclareFixedFont{\myfontb}{OT1}{ptm}{bx}{n}{8pt}
\usepackage{adjustbox}

\usepackage{thmtools}
\usepackage{thm-restate}

\usepackage{marvosym}
\makeatletter
\def\@fnsymbol#1{\ifcase#1\or \text{\Letter}\or *\or \dagger\or \ddagger\else\@arabic{#1}\fi}
\makeatother

\usepackage{lastpage}
\jmlrheading{23}{\ Preprint 2025}{1-\pageref{LastPage}}{1/21; Revised 5/22}{9/22}{21-0000}{}

\firstpageno{1}

\begin{document}

\title{Enhance Learning Efficiency of Oblique Decision Tree via Feature Concatenation}

\author{\name Shen-Huan Lyu \email lvsh@hhu.edu.cn \\
       \addr Key Laboratory of Water Big Data Technology of Ministry of Water Resources,\\ College of Computer Science and Software Engineering, Hohai University, Nanjing, China\\
       \addr State Key Laboratory for Novel Software Technology,
       Nanjing University, Nanjing, China\\
       \AND
       \name Yi-Xiao He~\thanks{Corresponding author} \email heyx@njucm.edu.cn \\
       \addr School of Artificial Intelligence and Information Technology, Nanjing University of Chinese Medicine, Nanjing, China
       \AND
       \name Yanyan Wang \email yanyan.wang@hhu.edu.cn \\
       \addr Key Laboratory of Water Big Data Technology of Ministry of Water Resources,\\ College of Computer Science and Software Engineering, Hohai University, Nanjing, China\\
       \AND
       \name Zhihao Qu \email quzhihao@hhu.edu.cn\\
       \addr Key Laboratory of Water Big Data Technology of Ministry of Water Resources,\\ College of Computer Science and Software Engineering, Hohai University, Nanjing, China\\
       \AND 
       \name Bin Tang \email cstb@hhu.edu.cn\\
       \addr Key Laboratory of Water Big Data Technology of Ministry of Water Resources,\\ College of Computer Science and Software Engineering, Hohai University, Nanjing, China\\
       \AND
       \name Baoliu Ye \email yebl@nju.edu.cn\\
       \addr State Key Laboratory for Novel Software Technology,
       Nanjing University, Nanjing, China
       }

\editor{My editor}

\maketitle

\begin{abstract}
    Oblique Decision Tree (ODT) separates the feature space by linear projections, as opposed to the conventional Decision Tree (DT) that forces axis-parallel splits. ODT has been proven to have a stronger representation ability than DT, as it provides a way to create shallower tree structures while still approximating complex decision boundaries. However, its learning efficiency is still insufficient, since the linear projections cannot be transmitted to the child nodes, resulting in a waste of model parameters. In this work, we propose an enhanced ODT method with Feature Concatenation (\texttt{FC-ODT}), which enables in-model feature transformation to transmit the projections along the decision paths. Theoretically, we prove that our method enjoys a faster consistency rate w.r.t. the tree depth, indicating that our method possesses a significant advantage in generalization performance, especially for shallow trees. Experiments show that \texttt{FC-ODT} can outperform the other state-of-the-art decision trees with a limited tree depth.
\end{abstract}


\section{Introduction}\label{sec:intr}

Tree-based ensemble methods, such as Random Forest \citep{breiman2001random,geurts2006extremely} and Gradient Boosting Decision Tree \citep{friedman2001greedy,chen2016xgboost,ke2017light}, have gained popularity in high-dimensional and ill-posed classification and regression tasks \citep{vershynin2018high}, for example on causal inference \citep{wager2018estimation,doubleday2022risk}, time series \citep{kane2014comparison}, or signal processing \citep{pal2005random}, but also for inference in applications such as image segmentation or object recognition in computer vision \citep{payet2012hough,kontschieder2014structured}. These methods are usually collections of decision trees with axis-aligned splits, such as CART~\citep[Section 2.2]{breiman1984classification} or C4.5~\citep{quinlan2014c4}, that is, the trees only split along feature dimensions, due to their computational efficiency and ease of tuning. 

However, the axis-parallel split methods often require very deep trees with complicated step-like decision boundaries when faced with high-dimensional data, leading to increased variance. Therefore, Olique Decision Tree (ODT) \citep[Section 5.2]{breiman1984classification} is proposed to use oblique decision boundaries, potentially simplifying the boundary structure. And axis-parallel decision trees can be considered a special case of ODT when the oblique projection direction is only selected from the set of basis vectors. Theoretically, ODT has been proven to have stronger representation capabilities and the potential to achieve better learning properties \citep{cattaneo2022convergence}. 

The major limitations of ODT are \textit{the excessive number of model parameters} in decision paths and \textit{high overfitting risk} in deep nodes \citep{cattaneo2022convergence}. In fact, the number of parameters required for the learning model to achieve a certain level of performance characterizes the \textbf{learning efficiency} of the algorithm. 
Although variant ODTs \citep{murthy1994system,carla1995multivariate} use linear combinations of features in each node, they do not transmit projection information to the child nodes. The way of retraining in each node wastes the model parameters we invested in the projection selection, leading to the insufficient learning efficiency of ODT
. On the other hand, as the tree grows, the number of samples in deep nodes rapidly decreases. ODT ignores the information obtained from the previous projection selection and retrains the linear model with limited samples, which can lead to severe overfitting risk \citep{shalev2014understanding}.


Previous studies often attempt to deal with these limitations via optimization but ignore the impact of wasted projection information. For example, \citet{zhu2020scalable} utilize the mixed-integer optimization (MIO) strategy to reduce the projection parameters by $\mathbb{L}^1$ regularization. 
\citet{lopez2013fisher} and \citet{tomita2020sparse} reduce the prediction variance of linear models by introducing $\mathbb{L}^2$ regularization terms or randomization, thus alleviating the overfitting risk. 
However, these methods can only deal with the learning process in each node separately and cannot consider the relationship between nodes in different layers. 
To overcome this challenge, we must note that, the transmission of projection information between nodes in different layers plays a critical role.

In this work, we propose an enhanced \texttt{O}blique \texttt{D}ecision \texttt{T}ree, \texttt{FC-ODT}, that leverages a \texttt{F}eature \texttt{C}oncatenation mechanism to improve learning efficiency. This mechanism facilitates a layer-by-layer feature transformation during the tree’s construction so that the optimized projection information in the node can be transmitted to its child nodes.
Meanwhile, the inductive bias brought by the concatenated features combined with the ridge regression method enables the retraining of the linear model in deep nodes to shrink high weights of original features across the linear model per the $\mathbb{L}^2$ penalty term. As a result, \texttt{FC-ODT} effectively mitigates multicollinearity by shrinking the coefficients of correlated features and helps alleviate overfitting by imposing a form of constraint.
The contributions of this work are summarized as follows:

\begin{itemize}[itemsep=0pt,topsep=0pt,leftmargin=20pt]
    \item We are the first to establish in-model feature transformation in a single decision tree, dealing with the problem of parameter waste caused by the split strategy of ODT.
    \item We prove that the consistency rate of \texttt{FC-ODT} is faster than traditional ODTs and demonstrate that the feature concatenation mechanism helps improve the learning efficiency of tree construction. 
    \item Experiments on simulated datasets verify the faster consistency rate of \texttt{FC-ODT}, and experiments on real-world datasets further show that \texttt{FC-ODT} outperforms other state-of-the-art ODTs.
\end{itemize}

\paragraph{Organization} The rest of this article is organized as follows. 
Section~\ref{sec:rlt} reviews some previous work. 
Section~\ref{sec:prl} introduces some essential background knowledge and notations. Section~\ref{sec:bodt} presents the \texttt{FC-ODT} for regression tasks. 
Section~\ref{sec:thr} proves the consistency rate of \texttt{FC-ODT}. 
Section~\ref{sec:proof} provides detailed proofs for the main theorem and lemma.
Section~\ref{sec:exp} conducts simulation and real-world experiments to verify our theoretical results. Section~\ref{sec:ccl} concludes our work with prospects.

\section{Related Work}\label{sec:rlt}

\subsection{Oblique Decision Tree}
ODT alleviates the problem of high variance of DT in high-dimensional settings, but faces extremely high complexity at each node to find the optimal split and suffers the overfitting risk at the deep node. To deal with these challenges, \citet{breiman1984classification} first use a fully deterministic hill-climbing algorithm to search for the best oblique split.
\citet{heath1993induction} and \citet{murthy1994system} propose combining random perturbations and hill-climbing algorithm to search for the best split, potentially avoiding getting stuck in local optima. 
Recently, \citet{bertsimas2017optimal} and \citet{zhu2020scalable} introduce the MIO strategy to further improve the efficiency of solving projection directions.
Unlike these deterministic approximation algorithms, another more interesting and practical research direction is to generate candidate projections through data-driven methods.
One possibility is to use dimensionality reduction techniques, such as PCA \citep{rodriguez2006rotation,menze2011oblique} and LDA \citep{li2003multivariate,lopez2013fisher}.
\citet{tomita2020sparse} show that sparse random projections or random rotations can also be introduced by incorporating.
Recently, some studies have extended ODTs to unsupervised learning frameworks such as clustering, demonstrating its advantages in representation ability~\citep{stepivsnik2021oblique,ganaie2022oblique}.
However, the explanation for their success is largely based on heuristics, until \citet{cattaneo2022convergence} demonstrate the consistency rate of excess risk for individual ODT.

\subsection{Feature Concatenation}
Deep Forest \citep{zhou2017deep} successfully constructs non-differentiable deep models by implementing feature concatenation mechanisms that enable in-model feature transformation based on decision trees. This mechanism has been theoretically proven to effectively improve the consistency rate of tree-based ensembles \citep{arnould2021analyzing,lyu2022depth}. \citet{chen2021improving} utilize the decision path of trees in the forest to generate oblique feature representations, which has been proven to effectively alleviate the risk of overfitting caused by feature redundancy \citep{lyu2022region}.
In addition, feature concatenation also has strong scalability and can adapt to different learning tasks by screening concatenated features. Recent research has expanded the tree-based deep models to some specific settings, such as multi-label learning \citep{yang2020multilabel} and semi-supervised learning \citep{wang2020learning}. 
Although feature concatenation has been widely used in ensemble learning, this work is still the first to introduce it in tree construction.

\section{Preliminary}\label{sec:prl}

In this work, we first describe the setting and notations related to tree-based estimators.

\subsection{Setting}  
We consider the regression setting, where the training set $S_n$ consists of $[0,1]^d\times\R$-valued independent random variables distributed as the prototype sample from a joint distribution $\mathbb{P}(\vx,y)=\mathbb{P}(\vx|y)\mathbb{P}(\vx)$ supported on $\mathcal{X}\times\mathcal{Y}$. This conditional distribution can be written as 
\begin{equation}
    y = f(\vx) + \epsilon\ ,
\end{equation}
where $f(\vx)=\E[y|\vx]$ is the conditional expectation of $y$ given $\vx$, and $\epsilon$ is a noise satisfying $\E[\epsilon]=0$ and $\Var[\epsilon]<\sigma^2$. The task considered in this work is to output a tree-based estimator $h_n(\cdot, T, S_n)\colon [0,1]^d\rightarrow\R$, where $T$ is a tree structure dependent on the training set $S_n$. To simplify notation, we denote $h_{T,n}(\vx)=h_n(\vx, T, S_n)$. The quality of a tree-based estimator $h_{T,n}$ is measured by its mean-of-squares error (MSE)
\begin{equation}\label{eq:mse}
    R(h_{T,n})=\E\left[(h_{T,n}(\vx)-f(\vx))^2\right]\ ,
\end{equation}
where the expectation is taken with respect to $\vx$, conditionally on $S_n$. As the training data size $n$ increases, we get a sequence of estimators $\{h_{T,i}\}_{i=1}^n$. A sequence of estimators $\{h_{T,n}\}_{n=1}^{\infty}$ is said to be \textit{consistent} if $R(h_{T,n})\to 0$ as $n\to \infty$.

\subsection{Trees} A decision tree is a data structure that is arranged and constructed in a top-down hierarchical manner using recursive binary partitioning in a greedy way. According to the CART methodology~\citep{breiman1984classification}, a parent node $t$ (representing a region in $\mathcal{X}$) within the tree is split into two child nodes, $t_L$ and $t_R$, to maximize the impurity decrease measured in MSE
\begin{equation}\label{eq:impurity}
\begin{adjustbox}{max width=\linewidth}
    $\widehat{\Delta}(b,\va,t)=\frac{1}{n}\sum_{\vx\in t}(y_i-\bar{y}_t)^2 - \frac{1}{n}\sum_{\vx\in t}\left(y_i - \bar{y}_{t_L}\1(\va^\top \vx_i\leq b) - \bar{y}_{t_R}\1(\va^\top \vx_i> b)\right)^2, $
\end{adjustbox}
\end{equation}
with respect to $(b,\va)$, with $\1(\cdot)$ denoting the indicator function and $\bar{y}_t$ denoting the sample average of the $y_i$ data whose corresponding $\vx_i$ data lies in the node $t$.

For traditional DT such as CART \citep[Section 2.2]{breiman1984classification}, splits always follow the direction parallel to the axis, which means that the projection direction $\va$ is limited to the set of standard basis vectors. In this case, the feature space partition learned by DT is always a set of hyper-rectangles. When encountering high-dimensional data, deep trees with high complexity are often generated, leading to overfitting risks. ODT such as oblique CART \citep[Section 5.2]{breiman1984classification} allows linear combinations between features as the basis for partitioning, thus expanding the hypothesis set of projection direction $\va$ to $\R^d$.

In Eqn.~\eqref{eq:mse}, the goal is to estimate the conditional mean response $f(\vx)$, the canonical tree output for $\vx\in t$ is $\bar{y}_t$, i.e., if $T$ is a decision tree, then
\begin{equation}\label{eq:tree_estimation}
    h_{T,n}(\vx)=\bar{y}_t=\frac{1}{n(t)}\sum_{\vx_i\in t}y_i\ ,
\end{equation}
where $n(t)$ denotes the number of samples in the node $t$.

The existing work to improve ODT mainly focuses on the computational efficiency and regularization of Eqn.~\eqref{eq:impurity}, which often only directly affects the splitting of nodes in each layer and cannot deal with the relationship between nodes in different layers. The average response in Eqn.~\eqref{eq:tree_estimation} wastes the computational cost invested in projection selection. The existing ODT learning frameworks are unable to effectively transmit the projection information of the parent node to its child nodes, resulting in limited learning properties.

\section{The Proposed Approach}\label{sec:bodt}

This section presents \texttt{FC-ODT}, whose key idea is to introduce a feature concatenation mechanism so that the layer-by-layer splitting process can achieve in-model feature transformation just like neural networks. 
\texttt{FC-ODT} can be not only practical but also a heuristic algorithm with provably better learning properties. 
It consists of three steps: feature concatenation, finding oblique splits, and tree construction, which are detailed in Algorithms~\ref{alg:feature_concatenation}-\ref{alg:bodt}.

\begin{figure}[ht]
    \centering
    \includegraphics[width=\linewidth]{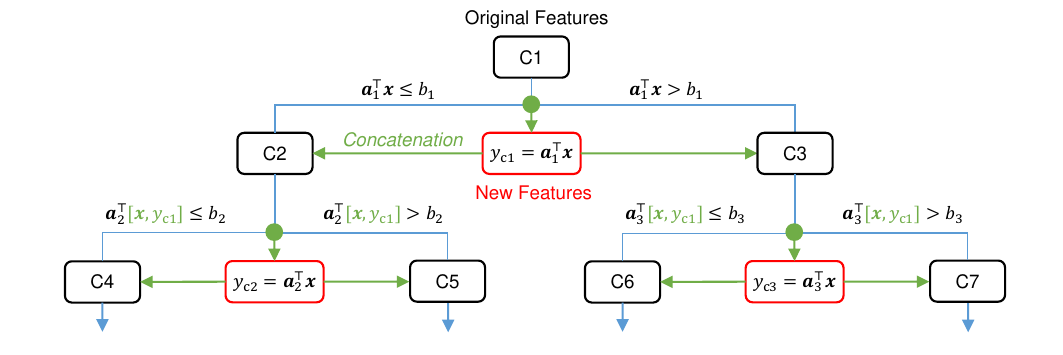}
    \vspace{-1em}
    \caption{Illustration of our \texttt{FC-ODT} framework, where $[x,y]$ denotes the feature concatenation between $x$ and $y$.}
    \label{fig:fc-odt}
\end{figure}

\begin{algorithm}[ht]
\caption{Feature concatenation.}
\label{alg:feature_concatenation}
\begin{spacing}{1.09}
\begin{algorithmic}[1]
\Require {The training data $\mX$ falling into the current internal node $t$ of a tree; $\tilde{\vy}_t$ denote the generated new feature for the training data of node $t$. 
}
\Ensure {The training data $\mX_t$ in node $t$.
}
\Function{$\mX_t \gets$ \textsc{FeaConc}$(\mX,\tilde{\vy}_t)$}{}
    \State{$\mX_{t} \gets [\mX,\tilde{\bm y}_t]\ $}
    \State{\Return $\mX_t$}
\EndFunction
\end{algorithmic}
\end{spacing}
\end{algorithm}
\begin{algorithm}[ht]
\caption{Finding the oblique split in an internal node.}
\label{alg:find_best_split}
\begin{algorithmic}[1]
\Require {The training set $S^{(t)}=(\mX,\vy)$ falling into the internal node $t$ of a decision tree; $\lambda$ is the hyper-parameter controlling the regularization strength.}
\Ensure {The projection $\va^*$ and the threshold $b^*$}
\Function{$(\va^*,b^*)\gets$ \textsc{FindObliqueSplit}$(S^{(t)})$}{}
    \State{$\va^*\gets \argmin_{\va}\|\vy - \mX^\top \va\|_2^2+\lambda\|\va\|_2^2$} 
    \State{$b^*\gets \argmax_{b} \widehat{\Delta}(b,\va^*,t)$} 
    \State{\Return $(\va^*,b^*)$}
\EndFunction
\end{algorithmic}
\end{algorithm}        

\subsection{Feature Concatenation}

As shown in Figure~\ref{fig:fc-odt}, when the parent node $t$ of each level splits, it will learn the oblique decision rule $\va^\top \vx<b$. We will record the projected score $\tilde{y}_t=\va^\top\vx$ as the augmented feature after entering the child node, and concatenate it with the original feature $\vx$ of the sample in the parent node to obtain the complete feature of the sample in the child node
\begin{equation}\label{eq:concatenation}
    \mX_{t} \gets [\mX
    ,\tilde{\bm y}_t]\ ,
\end{equation}
where $\mX_t\in [0,1]^{n\times d}$ denotes the feature matrix of samples in node $t$, and $[\cdot,\cdot]$ denotes the concatenation of two feature matrices to form a new feature matrix.
We encode the optimized projection information from the parent node into the new feature, and pass it along the decision path to the child nodes through concatenation operations to reduce the computational cost. This idea is similar to the Boosting algorithm \citep{bartlett1998boosting}, which uses different learners to process independent parts of data separately, and a stronger learner can be obtained by the ensemble. Feature concatenation has been used in deep forests to construct boosting frameworks to improve learning efficiency \citep{lyu2019refined}.
This feature concatenation process is repeated for each split until the desired tree size and constructs a \texttt{FC-ODT}, which is detailed in Algorithm~\ref{alg:feature_concatenation}.

\subsection{Finding Oblique Splits}

For \texttt{FC-ODT}, we rely on linear combinations of multiple features for binary splits at each node. For a sample $\vx\in\R^d$ in a feature space, the decision function at the node $t$ can be formulated as:
\begin{equation}
    \va_t^\top \vx < b_t
\end{equation}
with coefficients $\va_t$ denoting the projection direction for the split and splitting threshold $b_t$. Determining the optimal value for $\va_t$ is proved to be more challenging than identifying a single optimal feature and setting a suitable threshold for a split in the univariate scenario \citep{murthy1994system}. 

Considering that we have introduced a feature concatenation mechanism in the ODT generation process, this can lead to collinearity issues between the original and augmented features of the sample. Especially, the correlation between the augmented feature $\va^\top \vx$ and the label $y$ is high, which can cause overfitting problems during the training process as the augmented feature has a significant impact on the model's output. Therefore, we choose ridge regression to find the projection direction of the split
\begin{equation}
    \va_t(\lambda)=\argmax_{\va_t}\|\vy - \mX^\top \va_t\|_2^2+\lambda\|\va_t\|_2^2\ ,
\end{equation}
using regularization parameter $\lambda$. 

\begin{algorithm}[!ht]
\caption{Learning an oblique decision tree with feature concatenation (\texttt{FC-ODT}).}
\label{alg:bodt}
\begin{algorithmic}[1]
\Require {A training set $S_n=\{(\vx_1,y_1),\dots,(\vx_n,y_n)\}$; $\mX=[\vx_1,\dots,\vx_n]$ is the feature matrix; $\Gamma$ is a set of split eligibility criteria; pre-specified number of leaves $t_n$}
\Ensure {\texttt{FC-ODT} regression tree $T$}
\Function{$T \gets$ \textsc{TreeConstruction}$(S_n)$}{}
    \State{$t \gets 1$} \Comment{$t$ is the current node index}
    \State{$t_c \gets 1$} \Comment{$t_c$ is the current number of existing nodes}
    \State{$S^{(t)} \gets S_n$} \Comment{$S_n$ is the data falling into the root node}
    \While {$t < t_c+1$}
        \If{$\Gamma$ satisfied}
            \State{$(\va^*, b^*) \gets$ \textsc{FindObliqueSplit}$(S^{(t)})$} \Comment{refer to Algorithm~\ref{alg:find_best_split}}
            \State{$\Tilde{\bm y}_t\gets \mX_t^\top \va^*$} \Comment{generate the new feature via linear projection}
            \State{$\mX_t \gets \textsc{FeaConc}(\mX,\tilde{\vy}_t)$} \Comment{update $\mX_t$, refer to Algorithm~\ref{alg:feature_concatenation}}
            \State{$\vy_t \gets \vy_t-\Tilde{\bm y}_t$} \Comment{update $\vy_t$}
            \State{$S^{(K+1)} \gets \{i\colon \vx_i^\top \va^* < b^*,  \quad \forall i \in S^{(t)}\}$} \Comment{assign data to left child node}
            \State{$S^{(K+2)} \gets \{i\colon \vx_i^\top \va^* \geq b^*,  \quad \forall i \in S^{(t)}\}$} \Comment{assign data to right child node}
            \State{$\va_t\gets \va^{*}$} \Comment{store the projection vector for the current node}
            \State{$b_t\gets b^*$} \Comment{store the split threshold for the current node}
            \State{$\kappa_t\gets \{t_c+1,t_c+2\}$} \Comment{node indices of children of the current node}
            \State{$t_c=t_c+2$} \Comment{update the number of existing nodes}
        \Else 
            \State{$(\va_t,b_t,\kappa_t)\gets$ NULL}
        \EndIf
        \State{$t\gets t+1$} \Comment{move to the next node}
    \EndWhile
    \State{\Return $\left(S^{(1)},\{\va_t,b_t,\kappa_t,\tilde{\bm y}_t\}_{t=1}^{t_n}\right)$}
\EndFunction
\end{algorithmic}
\end{algorithm}

With this choice, the node model is optimized as \citep{smith1980critique}
\begin{equation}
    \va_t(\lambda)=\argmax_{\|\va_t\|=1} \Cov(\mX^\top\va_t,\vy)\cdot \frac{\Var(\mX^\top\va_t)}{\Var(\mX^\top\va_t)+\lambda}\ .
\end{equation}

Ridge regression can shrink high feature weights across the linear model per the $\mathbb{L}^2$ penalty term. This reduces the complexity of the model and helps make model predictions less erratically dependent on any one or more features.

\subsection{Tree Construction}

With these two steps at hand, we can iteratively perform node splitting and feature concatenation to construct an enhanced oblique decision tree with the ability of in-model feature transformation, named \texttt{FC-ODT}. The tree construction procedure is described in detail in Algorithm~\ref{alg:bodt}. 

To prepare for analyzing the convergence rate of excess risk in Section~\ref{sec:thr}, here we introduce the following definition.

\begin{definition}[Orthonormal decision stumps]\label{def:orthonormal}
The orthonormal decision stumps is defined as
\begin{equation}\label{eq:orthonormal}
    {\bm \psi}_t(\vx)=\frac{\vc_{t_L}(\vx)n(t_R) - \vc_{t_R}(\vx)n(t_L)}{\sqrt{w(t)n(t_L)n(t_R)}}\ ,
\end{equation}
where $\vc_{t_L}(\vx)=\vx^\top \left(\mX_{t_L}^\top \mX_{t_L} + \lambda \mI\right)^{-1}\mX_{t_L}^\top$ and $\mX_{t_L}=\mX \circ \1(\vx\in t_L)$, where $\circ$ is an elements-wise product, for internal nodes $t\in[T]$, where $w(t) = n(t)/n$ denotes the proportion of observations that are in $t$.
\end{definition}

The decision stump $\psi_t$ in Definition~\ref{def:orthonormal} is produced from the Gram–Schmidt orthonormalization of the projection functions $\{\vc_t(\vx),\vc_{t_L}(\vx)\}$ with respect to the empirical inner product space.
Next, we use a lemma to demonstrate that tree estimation is equal to the empirical orthogonal projection of $\vy$ onto the linear span of orthonormal decision stumps.

\begin{restatable}[Orthogonal tree expansion]{lemma}{orthogonal} \label{lem:orthogonal}
If $T$ denotes a decision tree constructed by \texttt{FC-ODT} method, then its output \eqref{eq:tree_estimation} admits the following orthogonal expansion
\begin{equation} 
h_{T,n}(\vx)=\sum_{t\in [T]}\langle\vy,{\bm \psi}_t\rangle_n \cdot\psi_t(\vx)\ ,
\end{equation}
where ${\bm \psi}_t=(\psi_t(\vx_1),\dots,\psi_t(\vx_n))^\top$ is defined in Definition~\ref{def:orthonormal}.

By construction, $\|{\bm \psi}_t\|=1$ and $\langle{\bm \psi}_t,{\bm \psi}_{t'}\rangle=0$ are satisfied for distinct internal nodes $t$ and $t'$ in $[T]$. In other words, $h_{T,n}$ is the empirical orthogonal projection of $\vy$ onto the linear span of $\{{\bm \psi}_t\}_{t\in [T]}$. Furthermore, we have
\begin{equation}
    |\langle\vy,{\bm \psi}_t\rangle|^2=\widehat{\Delta}(\hat{b},\hat{\va},t)\ .
\end{equation}
\end{restatable}

\begin{remark}
    Unlike the orthogonal decision stumps of conventional ODT, which iteratively projects the data onto the space of all constant predictors within a greedily obtained node \citep{cattaneo2022convergence}, Lemma~\ref{lem:orthogonal} shows that \texttt{FC-ODT} uses a feature concatenation mechanism to make the prediction factors of its orthogonal decision stumps contain information about the projection selection. Especially, the inductive bias brought by new features leads to child nodes tending to search for residual ridge regression solutions near the projection of the parent node, which enables the child nodes to learn local linear structures more efficiently in the subspaces.
\end{remark}

\section{Theoretical Analysis}\label{sec:thr}
In this section, we show that \texttt{FC-ODT} can achieve a faster convergence rate of consistency with respect to the tree depth $K$.

We consider an additive regression model to satisfy the following definition:
\begin{definition}[Ridge expansions  \citep{cattaneo2022convergence}]\label{def:ridge_expansions}
    Consider the family of functions consisting of finite linear combinations of ridge functions:
    \begin{equation}\label{eq:assumption}
    \begin{adjustbox}{max width=\linewidth}
    $\mathcal{G}=\left\{g(\vx)=\sum_{k=1}^{K}g_k(\va_k^\top \vx),\ \va_k\in \mathbb{R}^d, g_k\colon \mathbb{R}\rightarrow \mathbb{R}, \right.
    \left.\ k=1,\dots,K,\ \|g\|_{\mathcal{L}_1}<\infty \right\},$  
    \end{adjustbox}
    \end{equation}
    where $\|g\|_{\mathcal{L}_1}$ is a total variation norm defined in Definition~\ref{def:total_variation}.
\end{definition}

\begin{definition}[Total variation norm in node $t$]\label{def:total_variation}
    Define the total variation of a ridge function $\vx \rightarrow h(\va^\top \vx)$ in the node $t$ as $V(h,\va,t)=\sup_{\mathcal{P}}\sum_{\ell=0}^{|\mathcal{P}|-1}|h(z_{\ell+1}-h(z_{\ell}))|$, where the supremum is over all partitions $\mathcal{P}=\{z_0,\dots,z_{|\mathcal{P}|}\}$ of the interval $I(\va,t)=[\min_{\vx\in t}\va^\top\vx,\max_{\vx\in t}\va^\top\vx]$. For any $f\in \mathcal{F}=\operatorname{cl}(\mathcal{G})$, we define the $\mathcal{L}_1$ total variation norm of $f$ in the node $t$ as
    \begin{equation}
    \begin{adjustbox}{max width=\linewidth}
        $\|f\|_{\mathcal{L}_1(t)} \triangleq \liminf_{\epsilon \downarrow 0\ g\in \mathcal{G}} \left\{\sum_{k=1}^{K}V(g_k,\va_k,t)\colon g(\vx)=\sum_{k=1}^{K}g_k(\va_k^\top \vx),\ \|f-g\|\leq\epsilon\right\}.$
    \end{adjustbox}
    \end{equation}
\end{definition}

The models that decompose the regression function into a sum of ridge functions have been widely recognized and promoted by \citet{stone1985additive}, as well as \citet{hastie1987generalized}. In particular, the consistency of ODT under this assumption has been proven by \citet{cattaneo2022convergence}. On this basis, we study the impact of the feature concatenation in \texttt{FC-ODT}. Our results rely on the following assumption regarding the data-generating process.

\begin{restatable}[Exponential tails~\citep{cattaneo2022convergence}]{assumption}{exponential tails} \label{asp:exponential_tail}
The conditional distribution of $y$ given $\vx$ has exponentially decaying tails. That is, there exist positive constants $c_1, c_2$, and $M$, such that for all $\vx\in \mathcal{X}$, 
\begin{equation}
    \mathbb{P}(|y|>B+M \left| \ \vx\right.)\leq c_1\exp(-c_2B^2),\  B\geq 0\ .
\end{equation}
\end{restatable}

\begin{restatable}[Consistency rate for \texttt{FC-ODT}]{theorem}{consistency} \label{thm:consistency}
Let the conditional expectation $f(\vx)$ be from the ridge functions defined by 
Definition~\ref{def:ridge_expansions} and the conditional distribution $\mathbb{P}(y|\vx)$ satisfying
Assumption~\ref{asp:exponential_tail}. 
Consider a training set of $n$ samples drawn from this distribution and a $K$-layer decision tree $T_{K}$ constructed by \texttt{FC-ODT} on the training set.
Then, for any $K\geq 1$ and $n \geq 1$, we have
\begin{equation}
\begin{adjustbox}{max width=\linewidth}
    $\E\left[\|h_{T_K,n}(\vx)-f(\vx)\|_2^2\right]
    \leq 2 \inf_{f\in \mathcal{F}}\left\{\|g-f\|_2^2 + C_1\frac{\|g\|_{\mathcal{L}_1}^2}{K^2} + C_2\frac{2^Kd\log^2n}{n}\right\},$
\end{adjustbox}
\end{equation}
where $C_1=C_1(B,M)$ and $C_2=C_2(c_1.c_2,B,M)$ are two positive constants.
\end{restatable} 

\begin{remark}
Theorem~\ref{thm:consistency} proves that \texttt{FC-ODT} is consistent, and the convergence rate is the same as conventional ODTs with the increase of sample size $n$, both of which are better than axis-parallel DT. Since the feature concatenation mechanism transmits the projection information to child nodes, the convergence rate of excess risk is $\mathcal{O}(1/K^2)$ \textit{w.r.t.} tree depth $K$, which is faster than conventional ODTs with order $\mathcal{O}(1/K)$ in \citep{cattaneo2022convergence}. This result demonstrates that \texttt{FC-ODT} has advantages in learning efficiency compared to conventional ODTs. Especially considering the high computational complexity of optimal linear projection and the fact that deep decision paths can impair the interpretability of the model, we often limit the tree depth of ODT in practice~\citep{zhu2020scalable}. When $K$ is small, the theoretical advantages of \texttt{FC-ODT} over ODT become more significant.
\end{remark}

\section{Proofs}\label{sec:proof}

In this section, we provide the detailed proofs for the main theorem and lemma.

\subsection{Proof of Lemma~\ref{lem:orthogonal}}
\begin{proof}
Set $\mathcal{U}_t=\{u(\vx)\1(\vx\in t_L)+v(\vx)\1(\vx\in t_R): u,v\in \operatorname{span}(\mathcal{H})\}$ and consider the closed subspace $\mathcal{V}_t=\{v(\vx)\1(\vx\in t)\colon v\in \operatorname{span}(\mathcal{H})\}$. By the orthogonal decomposition property of Hilbert spaces, we can express $\mathcal{U}_t$ as the direct sum $\mathcal{V}_t \oplus \mathcal{V}_t^\perp$, where $\mathcal{V}_t^\perp=\{u\in \mathcal{U}_t\colon \langle u,v\rangle_n=0, \forall v\in \mathcal{V}_t\}$. Let $\Psi_t$ be any orthonormal basis for $\mathcal{V}_t$ that includes $w^{-1/2}(t)\1(\vx\in t)$, where $w(t)=n(t)/n$. Let $\Psi_t^\perp$ be any orthonormal basis for $\mathcal{V}_t^\perp$ that includes the decision stump defined by Eqn. \eqref{eq:orthonormal}.

Consider that $\tilde{y}_t(\vx)=\hat{\va}_t^\top \vx$ is the projection of $\vy$ onto $\mathcal{V}_t$, where $\hat{\va}_t=\left(\mX_{t_L}^\top \mX_{t_L} + \lambda \mI\right)^{-1}\mX_{t_L}^\top \vy$,
we have
\begin{equation}
    \tilde{y}_t(\vx)=\sum_{\psi\in\Psi_t}\langle\vy,{\bm \psi}\rangle_n \psi(\vx)\ ,
\end{equation}
and 
\begin{equation}
\begin{split}
    &\tilde{y}_{t_L}(\vx)\1(\vx\in t_L)+\tilde{y}_{t_R}(\vx)\1(\vx\in t_R)\\
    &=\sum_{\psi\in\Psi_t\cup\Psi_t^\perp}\langle\vy,{\bm \psi}\rangle_n \psi(\vx)\ .
\end{split}
\end{equation}
Using the above expansions, observe that for each internal node $t$,
\begin{equation}\label{eq:y_diff}
\begin{split}
    &\sum_{\psi\in\Psi_t^{\perp}}\langle\vy,{\bm \psi}\rangle_n \psi(\vx)
    = (\tilde{y}_{t_L}-\tilde{y}_{t})\1(\vx\in t_L)+(\tilde{y}_{t_R}-\tilde{y}_{t})\1(\vx\in t_R)\ .
\end{split}
\end{equation}

For each $\vx\in\mathcal{X}$, let $t_0,t_1,\dots,t_{K-1},t_{K}=t$ be the unique path from the root node $t_0$ to the terminal node $t$ that contains $\vx$. Next, sum \eqref{eq:y_diff} over all internal nodes and telescope the successive internal node outputs to obtain
\begin{equation}\label{eq:y_sum}
\begin{split}
    \sum_{k=0}^{K-1}(\tilde{y}_{t_{k+1}}(\vx)-\tilde{y}_{t_k}(\vx))&=\tilde{y}_{t_K}(\vx)-\tilde{y}_{t_0}(\vx)
    =\tilde{y}_{t}(\vx)-\tilde{y}(\vx)\ ,
\end{split}
\end{equation}
where $\tilde{y}$ is the linear estimation output by solving ridge regression in the root node:
\begin{equation}
    \min_{\va} \|\vy - \mX^\top \va\|_2^2+\lambda\|\va\|_2^2\ .
\end{equation}

Combining Eqns. \eqref{eq:y_diff} and \eqref{eq:y_sum}, we have
\begin{equation}
\begin{split}
\sum_{t\in [T]}\tilde{y}_t(\vx)\1(\vx\in t)&=\tilde{y} + \sum_{t\in [T]\backslash\{t_0\}}\sum_{\psi\in\Psi_t^\perp}\langle\vy,{\bm \psi}\rangle_n \psi(\vx)\\
&=\sum_{t\in [T]}\sum_{\psi\in\Psi_t^\perp}\langle\vy,{\bm \psi}\rangle_n \psi(\vx)\ ,
\end{split}
\end{equation}
where we recall that the root node $t_0$ is an internal node of $T$.

Finally, the decrease in impurity identity \eqref{eq:impurity} satisfies that:
\begin{equation}
\begin{adjustbox}{max width=0.97\linewidth}
$\begin{aligned}
\widehat{\Delta}(\hat{b},\hat{\va},t)
&=\frac{1}{n}\sum_{\vx_i\in t}(y_i-\tilde{y}_t(\vx))^2 \\
&\quad- \frac{1}{n}\sum_{\vx_i\in t} \left( y_i - \tilde{y}_{t_L}(\vx_i)\1(\vx_i\in t_L)- \tilde{y}_{t_R}(\vx_i)\1(\vx_i\in t_R) \right)^2\\
&=\left(\frac{1}{n}\sum_{\vx_i\in t}y_i^2-\sum_{\psi\in \Psi_t}|\langle\vy,{\bm \psi}\rangle_n|^2\right)
-\left(\frac{1}{n}\sum_{\vx_i\in t}y_i^2 -\sum_{\psi\in \Psi_t\cup\Psi_t^\perp}|\langle\vy,{\bm \psi}\rangle_n|^2\right)\\
&=\sum_{\psi\in\Psi_t^\perp}|\langle\vy,{\bm \psi}\rangle_n|^2 \ .
\end{aligned}$
\end{adjustbox}
\end{equation}

\end{proof}

\subsection{Proof of Theorem~\ref{thm:consistency}}
\begin{proof}
Following the proofs in \citep{cattaneo2022convergence}, we begin by splitting the MSE (averaging only with respect to the joint distribution of $\{\mathcal{A}_t\colon t\in [T_k]\}$) into two terms
\begin{equation}
    \E_{T_k} [\|h_{T_k,n}(\vx)-f(\vx)\|^2]=E_1+E_2\ ,
\end{equation}
where 
\begin{align}
    E_1=&\E_{T_k}[\|h_{T_k,n}(\vx)-f(\vx)\|^2]-2(\E_{T_k}[\|h_{T_k,n}(\vx)-\vy\|_n^2-\|\vy-f(\vx)\|_n^2])\notag\\
    &- \alpha(n,k) -\beta(n)\ ,\\
    E_2=&2(\E_{T_k}[\|h_{T_k,n}(\vx)-\vy\|_n^2]-\|\vy-f(\vx)\|_n^2)+\alpha(n,k) +\beta(n)\ ,
\end{align}
and $\alpha(n,k)$ and $\beta(n)$ are positive sequences that will be specified later.

To bound $\E[E_1]$, we split our analysis into two cases based on the observed data $y_i$. Accordingly, we have
\begin{equation}
    \E[E_1]=\E[E_1 \1(\forall i\colon |y_i|\leq B)]+\E[E_1 \1( i\colon |y_i|> B)]
\end{equation}

Firstly, we deal with the bounded term $\E[E_1 \1(\forall i\colon |y_i|\leq B)]$. According to \citep[pages 24-25]{cattaneo2022convergence} and \citep[Lemma 13.1, Theorem 11.4]{gyorfi2002distribution}, let $R=QB$ such that $R \geq M \geq \|\vy\|_{\infty}$, we have 
\begin{equation}\label{eq:appeq1}
\begin{split}
    &\mathbb{P}\left(\E_{T_K}[\|h_{T_k,n}(\vx)-f(\vx)\|^2]\geq E_2, \ \forall i\colon |y_i|<B\right)\\
    &\quad\leq 14 \sup_{\vx^n} \mathcal{N}\left(\frac{\beta(n)}{40R},\mathcal{F}_{n,k}(R),\mathcal{L}_1(\mathbb{P}_{\vx^n})\right)\exp\left(-\frac{\alpha(n,k)n}{2568R^4}\right)\ ,
\end{split}
\end{equation}
where $\mathcal{N}\left(\frac{\beta(n)}{40R},\mathcal{F}_{n,k}(R),\mathcal{L}_1(\mathbb{P}_{\vx^n})\right)$ denotes the covering number of $\mathcal{F}_{n,k}(R)$ by balls of radius $r>0$ in $\mathcal{L}_1(\mathbb{P}_{\vx^n})$ with respect to the empirical discrete measure $\mathbb{P}_{\vx^n}$ on $\vx^n$ which satisfies that
\begin{equation}\label{eq:appeq2}
    \mathcal{N}\left(\frac{\beta(n)}{40R},\mathcal{F}_{n,k}(R),\mathcal{L}_1(\mathbb{P}_{\vx^n})\right) \leq \left(3\left(\frac{enp}{d}\right)^d\right)^{2^k}\left(\frac{240eR^2}{\beta(n)}\right)^{\operatorname{VC}(\mathcal{H})2^{k+1}}\ .
\end{equation}

Combining Eqns.~\eqref{eq:appeq1} and \eqref{eq:appeq2}, we have
\begin{equation}
    \mathbb{P}(E_1\geq 0, \ \forall i\colon |y_i|\leq B)\leq 42 \left(\left(\frac{enp}{d}\right)^d\right)^{2^k}\left(\frac{240eR^2}{\beta(n)}\right)^{\operatorname{VC}(\mathcal{H})2^{k+1}}e^{-\frac{\alpha(n,k)n}{2568R^4}}\ ,
\end{equation}
so that $\mathbb{E}_1\geq 0, \forall i \colon |y_i|\leq B\leq 1/n^2$ and $E_1\1(\forall i\colon |y_i|\leq B)\leq 12R^2$.

Then, by choosing
\begin{equation}
\begin{adjustbox}{max width=\linewidth}
$\begin{aligned}
        \alpha(n,k)&= \frac{2568 R^4\left(2^k d \log (e n p / d)+2^k \log (3)+\mathrm{VC}(\mathcal{H}) 2^{k+1} \log \left(\frac{240 e R^2}{\beta(n)}\right)+\log \left(14 n^2\right)\right)}{n}\\
    \beta(n)&=\frac{240eR^2}{n^2}\ ,
\end{aligned}$
\end{adjustbox}
\end{equation}
we have 
\begin{equation}
    \E[E_1\1(\forall i\colon |y_i|\leq B)]\leq 12R^2\mathbb{P}(E_1\geq 0, \ \forall i\colon |y_i|\leq B)\leq \frac{12R^2}{n^2}\leq \frac{12Q^2B^2}{n^2}\ .
\end{equation}

Secondly, for the unbounded term $\E[E_1 \1( i\colon |y_i|> B)]$, by \citet{cattaneo2022convergence}, we have
\begin{equation}
\begin{split}
    &\E[\|h_{T_k,n}(\vx)-f(\vx)\|^2 \1( i\colon |y_i|> B)]\\
    \leq &(Q+1)^2\sqrt{(n+1)\mathbb{E}[y^4]}\sqrt{nc_1\exp(-c_2(B-M)^2)}\ .
\end{split} 
\end{equation}
and 
\begin{equation}
    \E[E_2]=\|f-g\|^2+\E[\|h_{T_k,n}(\vx)-\vy\|_n^2-\|\vy-g(\vx)\|_n^2]+\alpha(n,k)+\beta(n)\ .
\end{equation}

Since the excess risk can be decomposed by the Ridge expansions $g(\vx)$
\begin{equation}
\begin{adjustbox}{max width=\linewidth}
    $\E[\|h_{T_k,n}(\vx)-\vy\|_n^2-\|\vy-f(\vx)\|_n^2]=\|f-g\|^2+\E[\|h_{T_k,n}(\vx)-\vy\|_n^2-\|\vy-g(\vx)\|_n^2]\ ,$
\end{adjustbox}
\end{equation}

we have
\begin{equation}\label{eq:appeq5}
\begin{split}
    \E_{T_k} [\|h_{T_k,n}(\vx)-f(\vx)\|^2]\leq& \|f-g\|^2+\E[\|h_{T_k,n}(\vx)-\vy\|_n^2-\|\vy-g(\vx)\|_n^2]\\
    &+C_2\frac{2^K(d+\operatorname{VC}(\mathcal{H}))\log^2 n}{n}\ ,
\end{split}
\end{equation}
for some positive constant $C_2(c_1,c_2,B,M)$.

Define the squared node-wise norm and node-wise inner product as $\|\vf\|_t^2=\frac{1}{n(t)}\sum_{\vx_i\in t}(f(\vx_i))^2$ and $\langle\vf,\vg\rangle_t=\frac{1}{n(t)}f(\vx_i)g(\vx_i)$.We define the node-wise excess training error as 
\begin{equation}
    R_K(t)=\|h_{T_K,n}(\vx)-\vy\|_t^2-\|\vy-\vg\|_t^2\ .
\end{equation}   

Then, we define the total excess training error as:
\begin{equation}
    R_K=\sum_{t\in T_K} w(t)R_K(t), \quad w(t)=n(t)/n\ ,
\end{equation}
where $t\in T_K$ means $t$ is a terminal node of $T_K$.

According to the orthogonal decomposition of the \texttt{FC-ODT} in Lemma~\ref{lem:orthogonal}, we have
\begin{equation}\label{eq:recursive_app}
    R_K=R_{K-1}-\sum_{t\in T_{K-1}}\sum_{\psi\in \Psi_t^\perp}|\langle\vy, {\bm \psi}\rangle_n|^2\ .
\end{equation}

We denote by $\E_{T_K}$ the expectation is taken with respect to the joint distribution of $\{\mathcal{A}_t\colon t\in[T_K]\}$, conditional on the data. By the definition of $R_K$, we have
\begin{equation}
    \E[R_K]=\E[\|h_{T_K,n}(\vx)-\vy\|_n^2-\|\vy-\vg\|_n^2]\geq 0\ .
\end{equation}

Using the law of iterated expectations and the recursive relationship obtained in Eqn.~\eqref{eq:recursive_app}, we have
\begin{equation}\label{eq:appeq3}
\begin{split}
    \E_{T_K}[R_K]&=\E_{T_K}[\E_{T_K|T_{K-1}}[R_K]]\\
    &=\E_{T_{K-1}}[R_{K-1}]-\E_{T_{K-1}}\left[\E_{T_K|T_{K-1}}\left[\sum_{t\in T_{K-1}}\sum_{\psi\in \Psi_t^\perp}|\langle\vy,{\bm \psi}\rangle_n|^2\right]\right]\ .
\end{split}
\end{equation}

According to the sub-optimal probability defined by \citet[Section 2.2]{cattaneo2022convergence} and the sum of the iterative equality \eqref{eq:appeq3}, we have
\begin{equation}
    \E_{T_K}[R_K]=\sum_{t\in T_{K-1}\colon R_{K-1}(t)>0} P_{\mathcal{A}_t}(\kappa) \max_{(b,va)\in \R^{p+1}}\widehat{\Delta}(b,\va,t)\ ,
\end{equation}
where $P_{\mathcal{A}_t}(\kappa)=\mathbb{P}_{\mathcal{A}_t}\left(\max_{(b,\va)\in \mathbb{R}\times\mathcal{A}_t}\widehat{\Delta}(b,\va,t)\geq \kappa \max_{(b,\va)\in \mathbb{R}^{1+d}}\widehat{\Delta}(b,\va,t) \right)$ is defined to quantify the sub-optimality of the learning algorithm theoretically, and $\mathbb{P}_{\mathcal{A}_t}$ denotes the probability w.r.t. the randomness in the learning algorithm $\mathcal{A}_t$.

By \citep[Lemma 4 and Lemma 6]{cattaneo2022convergence}, we have
\begin{equation}
    \E_{T_K}[R_K]\leq \E_{T_{K-1}}[R_{K-1}]-\kappa\E_{T_{K-1}}\left[\frac{\E^2_{T_{K-1}}[R_{K-1}]/w(t)}{\sum_{t\in T_{K-1}}Q\|g\|_{\mathcal{L}_1(t)}^2}\right]\ .
\end{equation}

Due to the feature concatenation mechanism transmitting projection direction information within the model, the variation $\|g\|_{\mathcal{L}_1(t)}^2$ in the node $t$ in the path decreases with increasing depth. This phenomenon is similar to the decay of residuals in boosting algorithms, and we have obtained a recursion for $\E[R_K]$, by \citep[Lemma 5]{cattaneo2022convergence}, we have
\begin{equation}
    \E[R_K]\leq \frac{1}{\kappa\sum_{k=1}^{K}1/\E[\sum_{t\in T_{k-1}}w(t)Q\|g\|_{\mathcal{L}_1(t)}^2]}\ ,
\end{equation}
where
\begin{align}
    \E[\sum_{t\in T_{k-1}}w(t)Q\|g\|_{\mathcal{L}_1(t)}^2] &\leq \frac{\|g\|_{\mathcal{L}_1(t)}^2}{k-1}\E\left[\max_{t\in T_{k-1}}Q\sum_{t\in T_{k-1}}w(t)\right]\\
    & \leq \frac{\|g\|_{\mathcal{L}_1(t)}^2}{k-1} \E\left[\max_{t\in T_{K-1}}Q\right]\ .
\end{align}

Then, we obtain the inequality in the expected excess training error,
\begin{equation}\label{eq:appeq4}
    \E[R_K]\leq \frac{2Q\|g\|_{\mathcal{L}_1}^2}{ K(K-1)}\leq\frac{2M\|g\|_{\mathcal{L}_1}^2}{ BK(K-1)}\simeq\frac{2M\|g\|_{\mathcal{L}_1}^2}{ BK^2}\ .
\end{equation}

Finally, combining Eqns.~\eqref{eq:appeq4} and~\eqref{eq:appeq5}, we have
\begin{equation}
\begin{adjustbox}{max width=\linewidth}
    $\begin{aligned}
    \E\left[\|h_{T_K,n}(\vx)-f(\vx)\|_2^2\right]
    &\leq 2\|g-f\|_2^2+2\E[R_K]+C_2\frac{2^K\operatorname{VC}(\mathcal{H})\log^2n}{n}\\
    & \leq 2 \inf_{f\in \mathcal{F}}\left\{\|g-f\|_2^2 + C_1\frac{\|g\|_{\mathcal{L}_1}^2}{K^2} \right. \left. + C_2\frac{2^Kd\log^2n}{n}\right\}\ ,
\end{aligned}$
\end{adjustbox}
\end{equation}
for two positive constants $C_1(B,M)$ and $C_2(c_1,c_2,B,M)$.
\end{proof}

\section{Experiments}\label{sec:exp}

In this section, we verify our theoretical results on two simulated datasets that align with our assumptions. Moreover, we evaluate \texttt{FC-ODT} on eight real-world datasets and compare it to other state-of-the-art ODTs to demonstrate its superiority. 


\subsection{Results on Simulated Datasets}

\subsubsection{Implementation Detials}

\paragraph{Dataset} We generate two simulated datasets \textbf{sim1} and \textbf{sim2}. They are fully aligned with the assumption of our theoretical analysis, as stated in Eqn.~\eqref{eq:assumption}. 
All simulated datasets are generated as $y=f(\vx)+\epsilon$, where $\epsilon \sim \mathcal{N}\left(0,\sigma^2\right)$. 
For \textbf{sim1} dataset, 
\begin{equation}
\begin{split}
f(\vx)=&\operatorname{ReLU}(\vx_1) +\operatorname{ReLU}\left(\frac{\vx_2+\vx_3}{2}\right) \\
&+\operatorname{ReLU}\left(\frac{\vx_4+\vx_5+\vx_6}{3}\right) 
+\operatorname{ReLU}\left(\frac{\vx_7+\vx_8+\vx_9+\vx_{10}}{4}\right) \\
&+\operatorname{ReLU}\left(\frac{\vx_1+\vx_3+\vx_5+\vx_7+\vx_9}{5}\right)\ . 
\end{split}
\end{equation}
For \textbf{sim2} dataset,
\begin{equation}
\begin{split}
f(\vx)=&\exp(\vx_1) +\exp\left(\frac{\vx_2+\vx_3}{2}\right) \\
&+\exp\left(\frac{\vx_4+\vx_5+\vx_6}{3}\right) 
+\exp\left(\frac{\vx_7+\vx_8+\vx_9+\vx_{10}}{4}\right) \\
&+\exp\left(\frac{\vx_1+\vx_3+\vx_5+\vx_7+\vx_9}{5}\right)\ . 
\end{split}
\end{equation}
For each simulated dataset, $\vx$ is uniformly distributed over $[-3,3]^{10}$ and $\sigma=0.01$. Note that the test samples are generated without noise $\epsilon$, which aligns with the definition of consistency. 

\paragraph{Compared Methods} 
We choose a representative ODT method for comparison, i.e., \texttt{Ridge-ODT}~\citep{menze2011oblique}, which uses ridge regression to learn the optimal split direction. 
Since \texttt{FC-ODT} also uses ridge regression for the linear projection, \texttt{Ridge-ODT} can be seen as an ablation study without feature concatenation. On the other hand, the predictive performance varying with tree depth $K$ and number of sample $n$ for both methods can validate our theoretical findings.

\paragraph{Training Details} Experiments were run on a Windows 11 machine with a 3.40 GHz Intel i7-13700KF CPU and 32 GB memory. 
To prevent overfitting risk caused by insufficient sample size in leaf nodes, the minimum number of samples to split a node is set to $20$, and the minimum number of samples in leaf nodes is set to $8$. The regularization parameter $\lambda$ in \texttt{FC-ODT} and \texttt{Ridge-ODT} is searched for from the set $\{0.0001, 0.001, 0.01, 0.1, 1, 10, 100, 1000\}$ using a grid search with $5$-fold cross-validation on the training set.


\paragraph{Evaluation Protocol} The performance is measured by MSE (Mean Squared Error) on the test samples. The decrease in MSE reflects convergence. Each simulated dataset is randomly generated 10 times, and the average performance is reported.

\subsubsection{Convergence Rate \textit{w.r.t.} Tree Depth}\label{sec:exp-sim-depth}

\begin{figure}[t]
	\centering
	\subfigure[On sim1 dataset. \label{fig:fig1}]{
		\begin{minipage}[h]{0.48\linewidth}
			\centering
			\includegraphics[width=\textwidth]{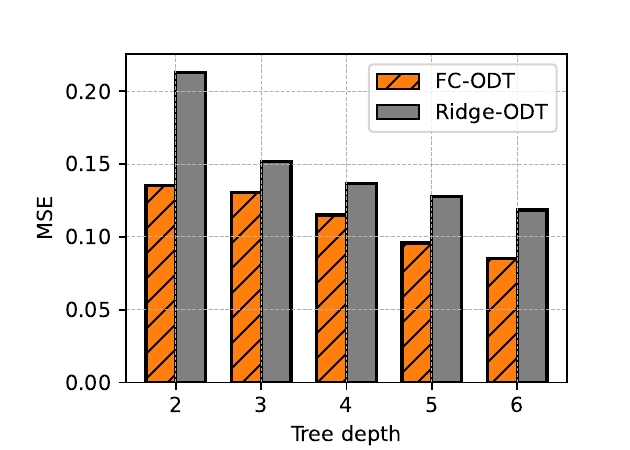}
	\end{minipage}}
    \hfill
	\subfigure[On sim2 dataset. \label{fig:fig2}]{
		\begin{minipage}[h]{0.48\linewidth}
			\centering
            \includegraphics[width=\textwidth]{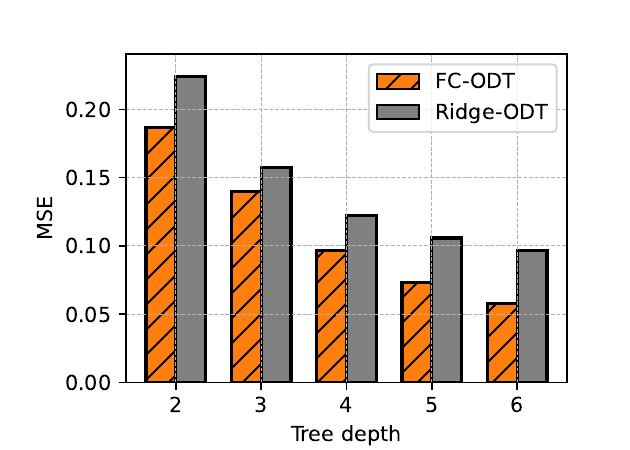}
	\end{minipage}}
	\caption{MSE values with different maximum tree depths.}
	\label{fig:depthconverge}
\end{figure}

This experiment aims to verify the theoretical advantage of \texttt{FC-ODT} on convergence rate w.r.t. the maximum tree depth. 
According to Theorem~\ref{thm:consistency}, when $n$ is sufficiently large, the $\mathcal{O}(1/n)$ term is sufficiently small, and the relationship between tree depth $K$ and test error can be studied. So we set the number of training samples to $2000$. 
Theorem~\ref{thm:consistency} indicates that the test error of \texttt{FC-ODT} should be considerably better than \texttt{Ridge-ODT} with a limited tree depth, so we set tree depth $K\in\{2,3,4,5,6\}$. We compare \texttt{FC-ODT} with \texttt{Ridge-ODT} on the two simulated datasets \textbf{sim1} and \textbf{sim2}, each time $2000$ training samples and $500$ test samples are randomly generated. 
Figure~\ref{fig:depthconverge} shows the average test MSE of \texttt{FC-ODT} and \texttt{Ridge-ODT} with the increasing tree depth. We can observe that \texttt{FC-ODT} has a faster convergence rate, which is consistent with our theoretical result. 

\begin{figure}[t]
	\centering
	\subfigure[On sim1 dataset. \label{fig:fig3}]{
		\begin{minipage}[h]{0.47\columnwidth}
			\centering
			\includegraphics[width=\textwidth]{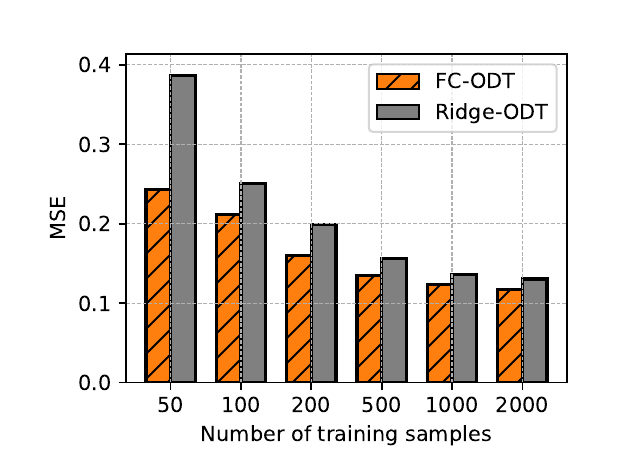}
	\end{minipage}}
    \hfill
	\subfigure[On sim2 dataset. \label{fig:fig4}]{
		\begin{minipage}[h]{0.47\columnwidth}
			\centering
			\includegraphics[width=\textwidth]{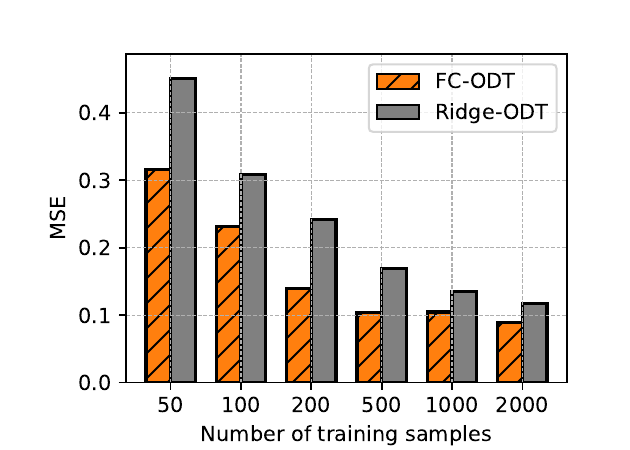}
	\end{minipage}}
	\caption{MSE values with different numbers of training samples.}
	\label{fig:nconverge}
\end{figure}

\subsubsection{Convergence Rate \textit{w.r.t.} Number of Samples}\label{sec:exp-sim-n}

Based on Figure~\ref{fig:depthconverge}, $K=4$ performs well. While deeper trees would result in a slight performance improvement, this comes at the cost of losing interpretability, which goes against the original intent of ODTs. Therefore, we choose $K=4$ as the regular setting thereafter. 
We compare \texttt{FC-ODT} and \texttt{Ridge-ODT} with increasing number of training samples $n\in\{50, 100, 200, 500, 1000, 2000\}$ on the two simulated datasets: \textbf{sim1} and \textbf{sim2}.
Figure~\ref{fig:nconverge} shows the test MSE of \texttt{FC-ODT} and \texttt{Ridge-ODT} with increasing number of samples. We can observe that both exhibit similar convergence tendency as the number of training samples increases, which is consistent with our theoretical result. 
However, \texttt{FC-ODT} always has a certain advantage in performance over \texttt{Ridge-ODT}, and its superiority is more pronounced when the number of training samples is limited. 
It suggests that the feature concatenation mechanism has potential for handling tasks with limited samples. This practical discovery is worth theoretical analysis in future work. 


\subsection{Results on Real-World Datasets}\label{sec:exp-real}

\subsubsection{Implementation Details}

\paragraph{Datasets}

We conduct experiments on two simulated datasets and eight LIBSVM regression datasets~\citep{chang2011libsvm}. Table~\ref{tab:datainfo} summarizes the detailed information of simulated and real-world datasets. Each dataset is randomly partitioned into training and test sets with ratio 3:2, and this partitioning process is repeated 10 times independently and the average result is reported. 

\begin{table}[t]
\centering
\caption{Dataset information. }
\begin{tabular}{l|cc}
\toprule
Dataset & N$_\text{samples}$ & N$_\text{features}$\\
\midrule
sim1 & 2000  & 10 \\
sim2 & 2000  & 10 \\
abalone & 4177  & 8 \\
bodyfat & 252  & 14 \\
cadata & 20640  & 8 \\
cpusmall & 8192  & 12 \\
housing & 506  & 13 \\
space\_ga & 3107  & 6 \\
mg & 1385  & 6 \\
mpg & 392  & 7 \\
\bottomrule
\end{tabular}
\label{tab:datainfo}
\end{table}

\paragraph{Compared Methods}
We test the $R^2$ of our method \texttt{FC-ODT} against \texttt{CART} and other state-of-the-art ODT methods. 
\begin{itemize}[itemsep=0pt,topsep=2pt,leftmargin=20pt]
    \item \texttt{S1O}~\citep{zhu2020scalable}: An ODT method using mixed-integer programs (MIP) to learn optimal split directions.
    \item \texttt{TAO}~\citep{carreira2018alternating}: An ODT method alternately updating the tree structure and linear combination to learn optimal split directions.
    \item \texttt{BUTIF}~\citep{barros2014framework}: An ODT method using embedded feature selection to learn optimal split directions.
    \item \texttt{Ridge-ODT}~\citep{menze2011oblique}: An ODT method using ridge regression to learn optimal split directions.
\end{itemize}

\paragraph{Training Details} 
This experiment's configurations are the same as the simulation experiment. Based on the simulation experiments, we keep the setting of tree depth $K=4$ for all the compared methods. 

\paragraph{Evaluation Protocol} 
Unlike the simulation experiments, in real-world problems, we do not know the underlying function $f(\vx)$. Therefore, in this experiment, we no longer compute MSE on the noise-free "test samples". Instead, we evaluate the performance on the randomly split test samples and use $R^2$ as the performance measure: 
\begin{equation}
    R^2=1-\frac{SS_{res}}{SS_{total}}\ ,
\end{equation}
where $SS_{res}$ is the sum of squared residuals, representing the difference between the predicted and actual values of the model,
and $SS_{total}$ is the total sum of squares, representing the difference between the actual value and the average value.
The value range of $R^2$ is between 0 and 1. The closer $R^2$ is to 1, the model explains more variance and fits better.

\subsubsection{$R^2$ Score Comparison} 

Table~\ref{tab:benchmark} reports the performance of the compared methods on ten datasets. The statistical significance of the experiments is tested through the Wilcoxon rank-sum test~\citep{rosner2003incorporation}. It shows that \texttt{FC-ODT} achieves the best average test $R^2$ score on eight out of ten datasets, and the best average rank compared to other state-of-the-art ODT methods: \texttt{Ridge-ODT}, \texttt{TAO}, \texttt{BUTIF} and \texttt{S1O}. It is significantly better than the baseline \texttt{CART} on all ten datasets. 
As we mentioned that \texttt{Ridge-ODT} can be viewed as an ablation study without feature concatenation, the comparison between \texttt{FC-ODT} and \texttt{Ridge-ODT} shows that the feature concatenation mechanism improves performance, indicating that transmitting projection in the decision paths improves the learning efficiency.

\begin{table}[ht]
\centering
\caption{Test $R^2$ (avg.$\pm$std. of $10$ times of running) on ten datasets. The best result is in bold. An entry is marked with a bullet `$\bullet$' (or `$\circ$') if \texttt{FC-ODT} is significantly better (or worse) than the corresponding method based on the Wilcoxon rank-sum test with confidence level $0.1$. }
\vspace{0.1in}
\resizebox{\linewidth}{!}{
\begin{tabular}{l|cccccc}
\toprule
Dataset & FC-ODT & Ridge-ODT & TAO & BUTIF & S1O & CART\\
\midrule
sim1 & \myfontb{0.877$\pm$0.007} & \myfont{0.849$\pm$0.010}$\bullet$ & \myfont{0.852$\pm$0.015}$\bullet$ & \myfont{0.772$\pm$0.013}$\bullet$ & \myfont{0.717$\pm$0.033}$\bullet$ & \myfont{0.588$\pm$0.023}$\bullet$\\
sim2 & \myfontb{0.895$\pm$0.019} & \myfont{0.874$\pm$0.011}$\bullet$ & \myfont{0.886$\pm$0.009}$\bullet$ & \myfont{0.799$\pm$0.010}$\bullet$ & \myfont{0.772$\pm$0.017}$\bullet$ & \myfont{0.724$\pm$0.018}$\bullet$\\
abalone & \myfont{0.541$\pm$0.021} & \myfontb{0.548$\pm$0.010} & \myfont{0.473$\pm$0.015}$\bullet$ & \myfont{0.417$\pm$0.028}$\bullet$ & \myfont{0.467$\pm$0.020}$\bullet$ & \myfont{0.447$\pm$0.012}$\bullet$\\
bodyfat & \myfontb{0.955$\pm$0.036} & \myfont{0.919$\pm$0.009}$\bullet$ & \myfont{0.875$\pm$0.032}$\bullet$ & \myfont{0.818$\pm$0.036}$\bullet$ & \myfont{0.940$\pm$0.024}$\bullet$ & \myfont{0.938$\pm$0.016}$\bullet$\\
cadata & \myfontb{0.701$\pm$0.004} & \myfont{0.700$\pm$0.005} & \myfont{0.671$\pm$0.008}$\bullet$ & \myfont{0.624$\pm$0.005}$\bullet$ & \myfont{0.552$\pm$0.012}$\bullet$ & \myfont{0.543$\pm$0.006}$\bullet$\\
cpusmall & \myfont{0.949$\pm$0.007} & \myfont{0.956$\pm$0.009}$\circ$ & \myfontb{0.962$\pm$0.004}$\circ$ & \myfont{0.935$\pm$0.005}$\bullet$ & \myfont{0.939$\pm$0.002}$\bullet$ & \myfont{0.934$\pm$0.011}$\bullet$\\
housing & \myfontb{0.776$\pm$0.029} & \myfont{0.764$\pm$0.031} & \myfont{0.752$\pm$0.031}$\bullet$ & \myfont{0.701$\pm$0.047}$\bullet$ & \myfont{0.717$\pm$0.060}$\bullet$ & \myfont{0.738$\pm$0.010}$\bullet$\\
space\_ga & \myfontb{0.583$\pm$0.035} & \myfont{0.562$\pm$0.013}$\bullet$ & \myfont{0.453$\pm$0.161}$\bullet$ & \myfont{0.478$\pm$0.021}$\bullet$ & \myfont{0.501$\pm$0.029}$\bullet$ & \myfont{0.508$\pm$0.018}$\bullet$\\
mg & \myfontb{0.657$\pm$0.016} & \myfont{0.640$\pm$0.018}$\bullet$ & \myfont{0.614$\pm$0.025}$\bullet$ & \myfont{0.536$\pm$0.021}$\bullet$ & \myfont{0.643$\pm$0.021}$\bullet$ & \myfont{0.625$\pm$0.021}$\bullet$\\
mpg & \myfontb{0.840$\pm$0.019} & \myfont{0.833$\pm$0.021} & \myfont{0.833$\pm$0.020} & \myfont{0.789$\pm$0.036}$\bullet$ & \myfont{0.805$\pm$0.039}$\bullet$ & \myfont{0.805$\pm$0.025}$\bullet$\\
\midrule
average rank & 1.30 & 2.40 & 3.30 & 5.20 & 4.10 & 4.70\\
\bottomrule
\end{tabular}
}
\label{tab:benchmark}
\end{table}

\begin{figure}[!ht]
    \centering
    \includegraphics[width=\linewidth]{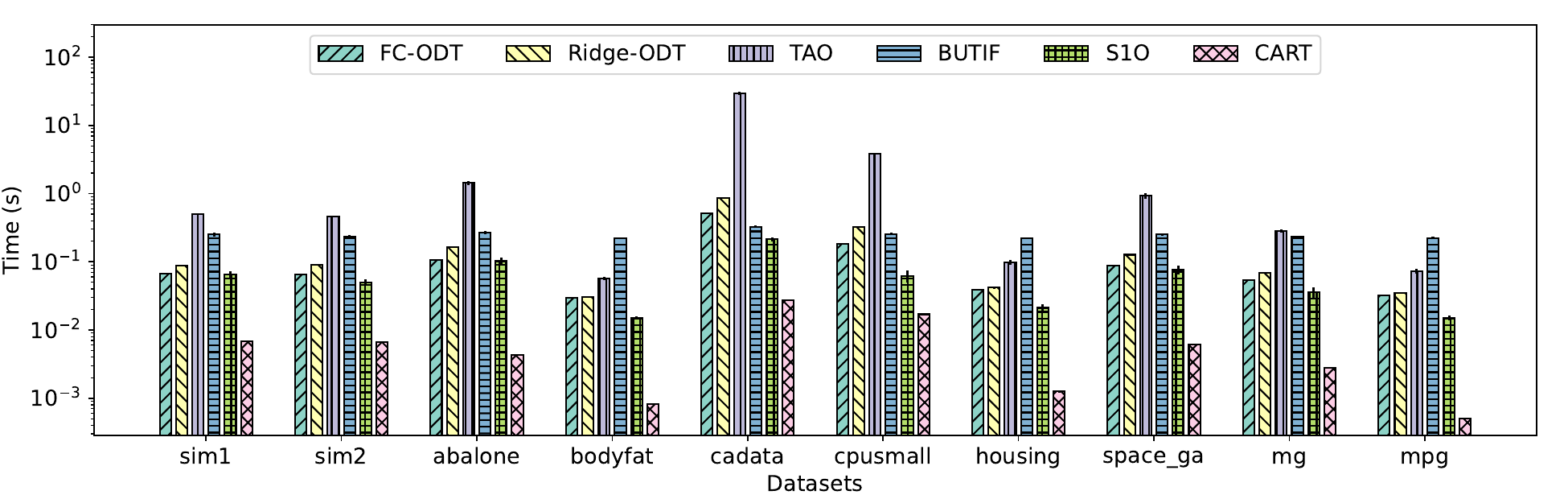}
    \caption{Running time on ten datasets. }
    \label{fig:timebar_tree}
\end{figure}

\subsubsection{Running Time Comparison} 
In Figure~\ref{fig:timebar_tree}, we plot the running time (mean $\pm$ std.) of all the compared methods. The running time includes the hyper-paramet
er tuning and the model training.
As we can observe in Figure~\ref{fig:timebar_tree}, the running time of \texttt{FC-ODT} is better than \texttt{TAO} and \texttt{BUTIF}, close to \texttt{Ridge-ODT} and \texttt{S1O}. It indicates that the feature concatenation mechanism hardly incurs any additional time overhead. 
However, the result shows that all the ODT methods have significantly longer running time than the axis-parallel tree, \texttt{CART}, mainly due to calculating the optimal linear projection.
Reducing the computational complexity of optimal linear projection in ODTs is still an open problem.

\section{Conclusion}\label{sec:ccl}

This work points out that the drawback of conventional ODTs lies in the waste of projection information during the tree construction. To address it, we propose \texttt{FC-ODT}, which introduces a feature concatenation mechanism to transmit the projection information of the parent node through in-model feature transformation, thereby enhancing the learning efficiency.
Both theory and experiments have verified that the projection information transmission brought by feature concatenation helps to improve the consistency rate. In future work, we will explore how to use random projection to enhance the diversity of concatenated features, thereby constructing an efficient random forest algorithm based on \texttt{FC-ODT}.


\acks{This work was supported by the National Science Foundation of China (62306104), The Hong Kong Scholars Program (XJ2024010), Natural Science Foundation of Jiangsu Province (BK20230949), China Postdoctoral Science Foundation (2023TQ0104), Jiangsu Excellent Postdoctoral Program (2023ZB140). }




\bibliography{sample}

\end{document}